\documentclass[10pt,bigtitle,twocolumn,SECTION,absheading]{article}
\usepackage{graphicx}
\usepackage{booktabs} 
\usepackage{amsfonts}
\usepackage{amssymb}
\usepackage{ragged2e}
\usepackage{bm}
\usepackage{mathtools}
\usepackage{amsmath}
\usepackage{supertabular,booktabs}
\usepackage{longtable}
\usepackage{ahsmeeting}
\usepackage{ragged2e}
\usepackage[format=plain,
            labelfont={bf,bf},
            textfont=bf]{caption}
\usepackage{array}
\usepackage{enumitem} 
\setlist[itemize]{noitemsep} 
\usepackage[colorlinks=true]{hyperref}
\usepackage[top=1.5cm,bottom=2cm,left=1cm,right=1cm,columnsep=11pt]{geometry} 
\newcounter{isorefi}
\def\isolatedrefs{\list{\stepcounter{isorefi}\textsuperscript{\arabic{enumi}}}{
          \settowidth\labelwidth{$^{9}$}
	  \leftmargin = 0pt
	  \listparindent = 0pt
	  \itemindent = \labelwidth
          \advance \itemindent by 4pt
	  \labelsep = 0pt
          \usecounter{enumi}\addtocounter{enumi}{\value{isorefi}}
        }
   \def\newblock{\hskip .11em plus .33em minus .07em}
   \sloppy
   \clubpenalty4000
   \widowpenalty4000
   \sfcode`\.=1000\relax
}

\begin{document}

\title{State-Space Identification of Unmanned Helicopter Dynamics using Invasive Weed Optimization Algorithm on Flight Data}

\author{\textbf{Navaneethkrishnan B}\\ Research Assistant \\ 
\and
\textbf{Pranjal Biswas}\\ Project Intern \\
\and
\textbf{Saumya Kumaar}\\  Research Assistant \\ 
\and
\textbf{Gautham Anand}\\  Research Assistant \\
\and
\textbf{S N Omkar}\\  Chief Research Scientist \\
\and 
Indian Institute of Science\\Bengaluru,INDIA\\
}

\date{}

\meeting{Presented at the 6th Asian/Australian Rotorcraft Forum / Heli Japan 2017, Kanazawa, Japan, November 7-9, 2017.\ Copyright~\copyright~2017 by the American Helicopter Society
International, Inc.\ All rights reserved.}

\abstract{In order to achieve a good level of autonomy in unmanned helicopters, an accurate replication of vehicle dynamics is required, which is achievable through precise mathematical modeling. This paper aims to identify a parametric state-space system for an unmanned helicopter to a good level of accuracy using Invasive Weed Optimization (IWO) algorithm. The flight data of Align TREX 550 flybarless helicopter is used in the identification process. The rigid-body dynamics of the helicopter is modeled in a state-space form that has 40 parameters, which serve as control variables for the IWO algorithm. The results after 1000 iterations were compared with the traditionally used Prediction Error Minimization (PEM) method and also with Genetic Algorithm (GA), which serve as references. Results show a better level of correlation of the actual and estimated responses of the system identified using IWO to that of PEM and GA.
 }

\maketitle

\section{Notation}

\begin{itemize}
\item A$_b$, B$_a$  : Flapping Angle Coupling Constants 
\item B$_d$, A$_c$  : Electronic Stabilizer bar-rotor coupling derivatives 
\item A$_{lat}$, A$_{lon}$, B$_{lat}$, B$_{lon}$ : Flapping Angle Control Derivatives 
\item C$_{lon}$, D$_{lat}$  : Intercoupling constants of the stabilizer bar
\item K$_r$, K$_{rfb}$  : Tail rotor damping coefficients
\item L$_u$, L$_v$, L$_b$, L$_w$   : Rolling Moment Derivatives
\item M$_u$, M$_v$, M$_w$, M$_a$, M$_{col}$  : Pitching Moment Derivatives
\item N$_v$, N$_p$, N$_w$, N$_r$, N$_{rfb}$, N$_{ped}$, N$_{col}$  : Directional Moment Derivatives
\item X$_u$, X$_a$  : Lateral Force Derivatives
\item Y$_v$, Y$_b$, Y$_{ped}$  : Longitudinal Force Derivatives
\item Z$_a$, Z$_b$, Z$_w$, Z$_r$, Z$_{col}$  : Heave Force Derivatives
\item \textit{a, b} : Lateral and longitudinal Flapping angle
\item \textit{c, d}: Lateral and longitudinal electronic stabilizer bar feedback
\item \textit{iter$_{max}$, iter} : Maximum and current iteration
\item \textit{n}: Nonlinear modulation index
\item \textit{p}, \textit{q}, \textit{r}: Roll, Pitch and Yaw Rate
\item \textit{r$_{fb}$}: Yaw rate feedback
\item \textit{u}, \textit{v}, \textit{w}: Lateral, longitudinal and vertical translational velocity
\item $\phi$, $\theta$: Roll and Pitch angles
\item $\delta_{lat}$, $\delta_{lon}$, $\delta_{ped}$, $\delta_{col}$ : Lateral, longitudinal, pedal and collective inputs
\item $\mu_A$, $\mu_B$ : Mean of sets A and B
\item $\sigma_A$, $\sigma_B$, $\sigma_{initial}$, $\sigma_{final}$, $\sigma_{iter}$: Standard deviations of sets A, B, the initial iteration, the final iteration and the current iteration respectively
\item $\tau_f$, $\tau_s$, : Bare rotor and stabilizer bar time constants
\end{itemize}

\section{Introduction}

  The use of unmanned helicopters have been increasing steadily in the past decade not only for military applications but also in civilian applications like crop-dusting, seed-bombing, search and rescue etc. (Ref. 1-4). The popularity stems from the vertical take-off and landing (VTOL) capabilities of helicopters.  For a given amount of battery power, helicopters provide longer service range and hence can cover larger area when compared to commonly used quad-copters (Ref. 5) However, autonomous operation is favorable to improve the efficiency of the task or even render it feasible (for example, in a case where the vehicle needs to fly out of communication range of the radio transmitter). Design of controllers to enable autonomy in unmanned helicopters proves to be an arduous task due to the complexity involved in obtaining the mathematical model of the same.  This difficulty in modeling arises due to the non-linear dynamics of helicopters and a high degree of inter axis coupling. 
\begin{figure}[h]
\includegraphics[width=0.48\textwidth]{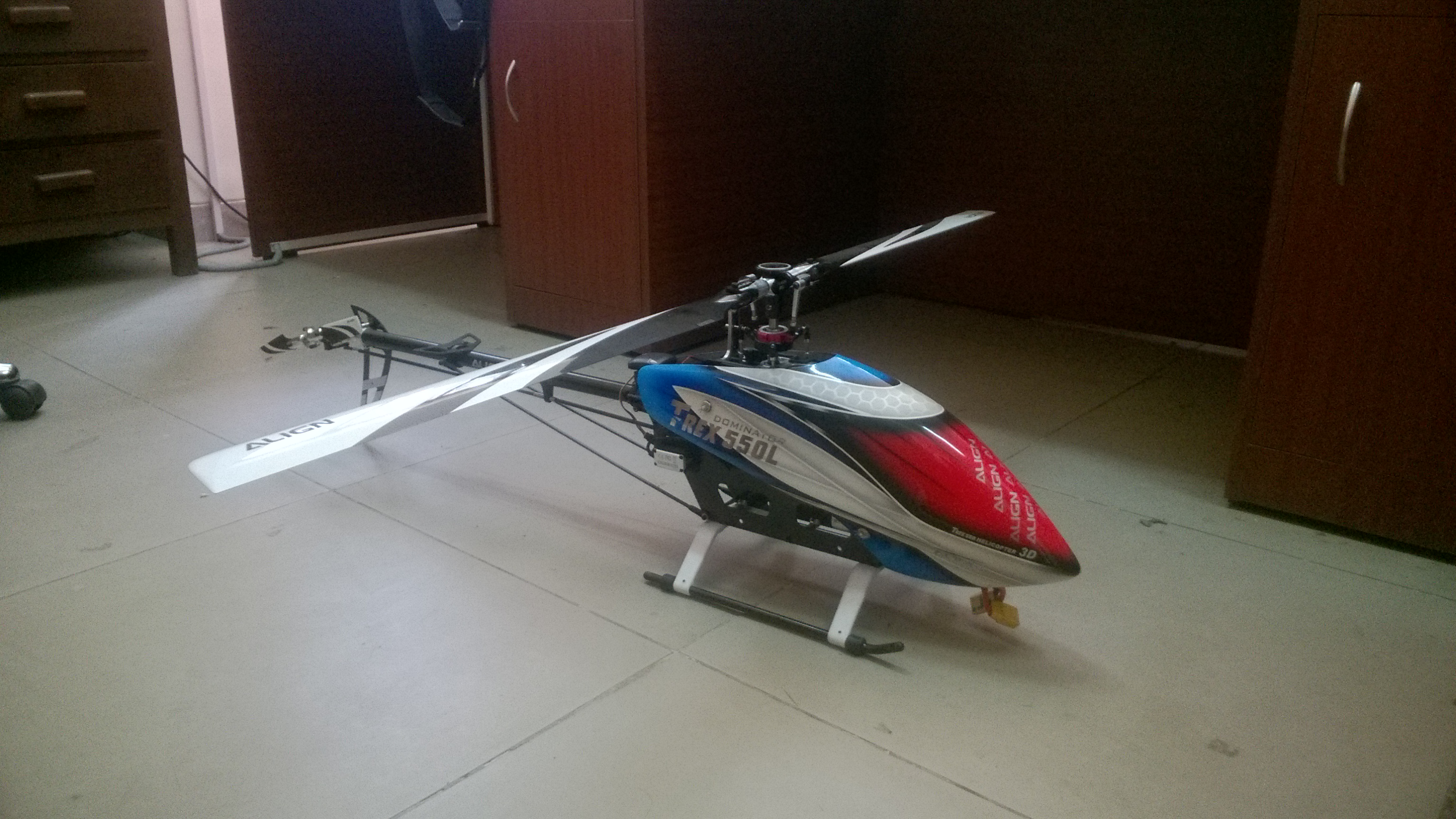}
\caption{Align TREX 550L - Helicopter used for data collection}
\label{fig:figure2}
\end{figure}
There have been many attempts at modeling both small scaled and its full sized counterpart. One of the most notable pieces of research on this topic was carried out by Mettler \textit{et al.} (Ref. 6-8). The research uses a frequency domain identification tool called CIFER (Comprehensive Identification from Frequency Responses), which was developed by the U.S Army and NASA, to identify the state-space parameters of a Yamaha R-50 helicopter. However, due to the high price and limited applicability of the CIFER® tool, we decided to use an optimization algorithm to estimate the parameters of the state-space model. 

Many studies have been conducted evaluating the effectiveness of optimizing algorithms like Prediction Error Minimization (PEM) (Ref. 9), Subspace Method (Ref. 10) and Recursive Least squares (Ref.11, 12) in identifying dynamic systems. However, these traditionally used algorithms work well only for simple linear dynamic models because of their tendency to get stuck in local optima. An exhaustive study has been conducted on the potency of meta-heuristic algorithms on global optimization problems and has been presented in (Ref. 13). The unassuming nature of the algorithms towards the differentiability of the cost-function and their dynamic nature enable it to jump out of any local minima. The effectiveness of meta-heuristic algorithm in identifying the yaw-heave dynamics of a small scaled helicopter has been demonstrated by Tang et. al. (Ref 14).

Invasive Weed Optimization (IWO) algorithm is a population based random search algorithm which was proposed by Mehrabian et.al. (Ref. 15) inspired by the colonizing behavior of invasive weed in agricultural farmlands. The algorithm is structurally simple, robust and has lesser hyper-parameters as compared to its counterparts. The algorithm has found use in a wide range of application such as Neural Network training (Ref. 16), combustion optimization (Ref. 17), PID controller design (Ref 18), trajectory planning (Ref. 19), auto-disturbance rejection controller tuning (Ref. 20) etc.

This paper aims to obtain a mathematical model of an unmanned helicopter (Fig 1.) in the state-space form using IWO algorithm and compare its effectiveness against the traditionally used PEM and a commonly used meta-heuristic tool like Genetic Algorithm (GA). 
The following section describes the physical characteristics and the onboard instrumentation of the helicopter used to collect the flight data which is followed by the methodology used to acquire flight data including the flight test procedure. This is followed by the description of the state-space model used and then succeeded by the system identification methodology using the IWO algorithm. The paper concludes by tabulating the results and a discussion on the same.

\section{DESCRIPTION OF HELICOPTER UAV}

In order to be able to model a vehicle successfully, we need to record the outputs from the actual vehicle to be modeled.  The small-scaled helicopter used for modeling is an Align TREX 550L which is a hobby grade vehicle capable of 3-D flight. Unlike most commercially available hobby helicopters, the TREX 550L does not possess a flybar. A flybar is a mechanical device that, using the principles of Coriolis force, provides a negative feedback to the longitudinal and lateral motions. The negative feedback for the motions in the TREX 550L is provided by a Micro Electro-Mechanical System (MEMS) Gyroscope which serves as a replacement for the mechanical flybar. The advantage of using an electronic flybar is that the amplitude of the negative feedback can be tuned to the pilots comfort. It may be tuned to provide high stability or to provide high input sensitivity. Figure 2 describes the physical dimensions of the helicopter.

\begin{figure}[h]
\includegraphics[width=0.48\textwidth]{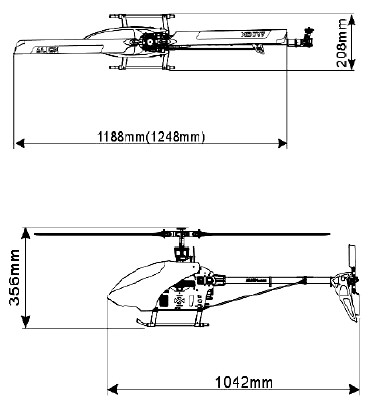}
\caption{Dimensions of the helicopter used for flight data acquisition}
\label{fig:figure2}
\end{figure}

\subsection{Helicopter Instrumentation}
The helicopter used is equipped with an array of state-of-the-art sensors and flight control board that record high-quality flight data. The flight control board used is Pixhawk, an ARM based flight controller. This arrangement records the velocities (\textit{u}, \textit{v} and \textit{w}), angular rates (\textit{p}, \textit{q}, and \textit{r}) and blade flapping angles (\textit{a} and \textit{b}, that are functions of swash-plate servo arm positions). The data is recorded onto a micro SD card on the Pixhawk at the rate of 100 Hz. The data is filtered for sensor noise using a Butterworth filter at half-power frequency. Figure 3 depicts the instrumentation mounted atop the unmanned helicopter.

\begin{figure}[h]
\includegraphics[width=0.48\textwidth]{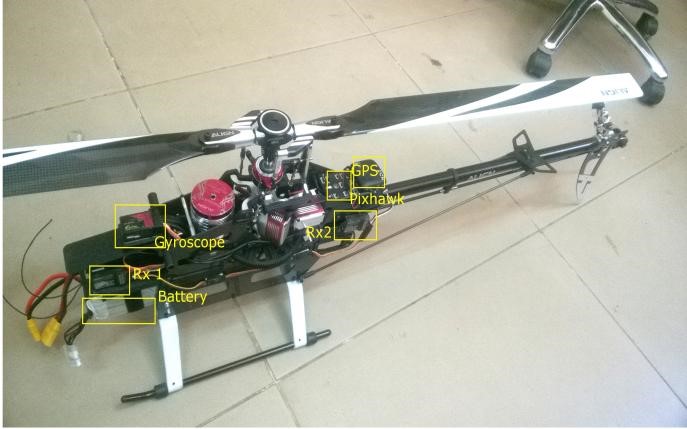}
\caption{Instrumentation of TREX 550L}
\label{fig:figure2}
\end{figure}

\section{FLIGHT DATA ACQUISITION}

A helicopter has a wide range of dynamic responses inclusive of responses in the nonlinear range. However, missions requiring autonomy rarely call for operations in the nonlinear domain. Hence, we linearize the system and model it in the state-space form. In order to obtain an accurate dynamic model, the maximum dynamic range of the vehicle should be excited while keeping it from entering the nonlinear domain. One of the well-established techniques to perform this excitation is the Frequency Sweep method (Ref.  21). In this method, the vehicle is given sinusoidal inputs of constant amplitude and   increasing frequency in both lateral and longitudinal direction. Frequency Sweep can be executed either using a signal-generator or manually.  To maintain the safety aspect of the flight test and considering the expertise of our pilot, we decided to manually perform the frequency sweep using the '3-2-1-1' input sequence. In a '3-2-1-1' sequence, a positive deflection of the control stick is made for the duration of 3 seconds and then a negative deflection is given for 2 seconds and the stick is deflected in the positive direction again for 1 second and then back to negative for 1 second. This sequence is repeated for both lateral and longitudinal directions. The pedal input was given only for directional corrections. The tests were conducted at constant rpm and hover conditions. In order to keep the external perturbations to the minimum, the flight tests were conducted during the period where ambient wind speed was lesser than 1.029 m/s (2 knots) as high winds can degrade the accuracy of the identified dynamic model.

\section{STATE SPACE MODEL}

The parametric state-space form of the unmanned helicopter was adopted from Mettler \textit{et al}. (Ref. 6), which has been derived from first principles. The state space form is defined as shown in Eq.1. Where x (Eq.2) is the state matrix and u (Eq.3) is the input matrix.

\begin{equation}
\dot{\textbf{x}} = A\textbf{x} + B\textbf{u}
\end{equation}

Where,
\begin{equation}
\textbf{x}= 
\begin{bmatrix}
    u       & v & p & q & \phi & \theta & a & b & w & r & r_{fb} & c & d
\end{bmatrix}^{T}
\end{equation}
\begin{equation}
u = 
\begin{bmatrix}
    \delta_{lat} & \delta_{lon} & \delta_{ped} & \delta_{col}
\end{bmatrix}^{T}
\end{equation}

\begin{equation}
A = 
\scriptsize{\begin{bmatrix}
    X_{u}&0&0&0&0&-g&X_{a}&0&0&0&0&0&0 \\
    0&Y_{v}&0&0&g&0&0&Y_{b}&0&0&0&0&0 \\
    L_{u}&L_{v}&0&0&0&0&0&L_{b}&L_{w}&0&0&0&0 \\
    M_{u}&M_{v}&0&0&0&0&M_{a}&0&M_{w}&0&0&0&0 \\
    0&0&1&0&0&0&0&0&0&0&0&0&0 \\
    0&0&0&1&0&0&0&0&0&0&0&0&0 \\
    0&0&0&-\tau_{f}&0&0&1&A_{b}&0&0&0&A_{c}&0 \\
    0&0&-\tau_{f}&0&0&0&B_{a}&-1&0&0&0&0&B_{d} \\
    0&0&0&0&0&0&Z_{a}&Z_{b}&Z_{w}&Z_{r}&0&0&0 \\
    0&N_{v}&N_{p}&0&0&0&0&0&N_{w}&N_{r}&N_{rfb}&0&0 \\
    0&0&0&0&0&0&0&0&0&K_{r}&K_{rfb}&0&0 \\
    0&0&0&-\tau_{s}&0&0&0&0&0&0&0&-1&0 \\
    0&0&-\tau_{s}&0&0&0&0&0&0&0&0&0&-1
\end{bmatrix}}
\end{equation}

\begin{equation}
B = 
\scriptsize{\begin{bmatrix}
    0&0&0&0 \\
    0&0&Y_{ped}&0 \\
    0&0&0&0 \\
    0&0&0&M_{col} \\
    0&0&0&0 \\
    0&0&0&0 \\
    A_{lat}&A_{lon}&0&0 \\
    B_{lat}&B_{lon}&0&0 \\
    0&0&0&Z_{col} \\
    0&0&N_{ped}&N_{col} \\
    0&0&0&0 \\
    0&C_{lon}&0&0 \\
    D_{lat}&0&0&0
\end{bmatrix}}
\end{equation}

The A and B matrices (Eq.4 and Eq.5) together have 40 parameters which require to be estimated for successful identification of the system dynamics.

\section{SYSTEM IDENTIFICATION}

The IWO algorithm is used to estimate these 40 parameters such that the model replicates the responses as close to the recorded training data as possible. The algorithm scatters seeds in a 40 dimensional search space and tries to converge at a global optimum. Which essentially means that the algorithm tries to find the best set of values for the state-space matrices. The steps involved in the identification is explained as follows.

\textit{Step 1 - Initialization}: A finite number of seeds (solutions) are deposited randomly over a search space. 

\textit{Step 2 - Cost Function}: Cost/Fitness of each seed is determined by running it through a cost function.The Population Pearson Coefficient is used to define a cost function for this algorithm. The Population Pearson correlation coefficient of two sets of data is a measure of their frequency correlation or rather their “shape fitness”. Consider data sets A and B have N scalar observations each, then the Pearson correlation coefficient is defined as shown in Eq.6 

\begin{equation}
\rho(A,B) = \frac{1}{N-1}\sum_{n=1}^{N}\Bigg(\frac{\overline{A_{l} - \mu_{a}}}{\sigma_{a}}\Bigg)\Bigg(\frac{\overline{B_{l} - \mu_{b}}}{\sigma_{b}}\Bigg)
\end{equation}

From this we define the cost function as depicted in Eq.7

\begin{equation}
C = \sum_{i=1}^{13} {(1 - \rho_{i})}^2
\end{equation}

\textit {Step 3 - Reproduction}: Every seed grows into a flowering plant which in turn deposits its own seeds. But the number of seeds each flowering plant can deposit linearly varies from the minimum possible seeds to the maximum possible seeds based on its fitness value. In other words, the plant with a higher fitness score can deposit the maximum number of seeds.

\textit {Step 4 - Spatial Dispersal}: The spawned seeds are distributed randomly over the search space by normally distributed random number with zero mean and varying variance. The variance varies as shown in Eq.8

\begin{equation}
\sigma_{iter} = \frac{(iter_{max} - iter)^n}{iter_{max}}(\sigma_{initial} - \sigma_{final}) + \sigma_{final}
\end{equation}

This ensures that the probability of dropping a seed in a distant area decreases nonlinearly with each iteration due to which fitter plants are grouped and the weak ones are eliminated.

\textit{Step 5 - Competitive Exclusion}: Initially all the seeds are allowed to grow unchecked till it reaches the population limit. Once the population limit is reached, a function to eliminate plants with poor fitness gets called. This function ranks all the seeds in the population with their parents'  ranks and eliminates weeds with lower fitness and allows newer seeds to fill the population. This way plants with low fitness scores are allowed to survive if their offspring gives a high fitness score.

The above steps are iterated 1000 times in order to obtain the highest fitness. The maximum iterations are set at 1000 because the cost plateaued out at around 1000 iterations during tuning. 10 trials of these iterations are conducted so that the confidence interval of 95\% can be calculated. A 95\% confidence interval means that the probability of the exact solution falling within that interval is 0.95 or 95\%.  The solution with the best fitness or the lowest cost is used to generate the state space model. The validation of this model is done using another dataset from the same helicopter.

\section{RESULTS}

Of the 13 identified states, we perform time domain verification only on \textit{p}, \textit{q}, $\phi$ and $\theta$ as they are the only states crucial for controller design. The results from the verification are tabulated in Table 1. Further the identified parameters with their 95\% confidence interval has been tabulated in Table 2.  It can be observed from Table 1 that PEM performs an inferior job of identifying the model as compared to the metaheuristic algorithms. This, as discussed in the introduction of this paper, is due to its inability to handle complex systems like helicopter dynamics due to its tendency to get stuck in a local optima. Among IWO and GA, we notice that IWO performs noticeably better than its counterpart. \\
We can observe that a few values in Table 2 are set 0 by the algorithm. Those parameters come into play on while in forward flight and not during hover. This corroborates the values which was identified using CIFER® by Mettler \textit{et al.} (Ref. 6) for hover condition.  Figures 4, 5 and 6 helps in visualizing the time domain verification performed for models identified using PEM, GA and IWO respectively.

\pagebreak
\captionof{table}{Performance Comparison} \label{tab:title} 
\centering
\begin{tabular}{cccc}
  \hline
  States & PEM & GA & IWO\\ 
  \hline
  p & 0.5535 & 0.8385 & 0.9575\\ 
  q & 0.5329 & 0.7691 & 0.9089\\ 
  $\phi$ & 0.4946 & 0.6765 & 0.9084\\ 
  $\theta$ & 0.0336 & 0.6576 & 0.8722\\ 
  \hline
  \end{tabular}

\captionof{table}{Parameters Identified using IWO} \label{tab:title}
\centering
\begin{tabular}[c]{cccc}
  \hline
  \multicolumn{4}{r}{95\% Confidence Interval} \\
  \cline{3-4}
  Parameter & Value & Upper Limit & Lower Limit\\ 
  \hline 
$X_{u}$ & -0.32066 & -0.212 & -0.32658\\
$X_{a}$ & 40.21598 & 40.43 & -5.36256\\
$Y_{v}$ & -0.93658 & -0.745 & -0.92882\\
$Y_{b}$ & -16.1151 & 18.05 &-28.9708\\
$L_{u}$ & -0.00121
& -0.0024
& -0.07221\\
$L_{v}$ & -0.47665
&-0.31
&-0.65863
\\
$L_{b}$ & 133.6111
&183.6
&145.5951
\\
$L_{w}$ & 0
&0
&0
\\
$M_{u}$ & 0.1
&0.10005
&0.099401
\\
$M_{v}$ &-0.09822
&-0.021
&-0.07774
\\
$M_{a}$ & 104.9063
&87.47161
&73.76749
\\
$M_{w}$ & 0
&0
&0
\\
$T_{f}$ & 0.093851
&0.176949
&0.09222
\\
$A_{b}$ & -0.19213
&-0.12113
&-0.43625
\\
$A_{c}$ & 0.061597
&0.351488
&0.057038
\\
$B_{a}$ & 0.083523
&0.125623
&-0.0954
\\
$B_{d}$ & 0.984168
&0.778003
&0.32217
\\
$Z_{a}$ & 8.166105
&8.446461
&-4.93675
\\
$Z_{b}$ & 1.028478
&8.586446
&-87.4542
\\
$Z_{w}$ & 0.045724
&0.038181
&0.017925
\\
$Z_{r}$ & -1.39101
&0.85415
&-0.28432
\\
$N_{v}$ & 0.009652
&0.008206
&0.003685
\\
$N_{p}$ & -8.23373
&-2.13475
&-6.88379
\\
$N_{w}$ & 0
&0
&0
\\
$N_{r}$ & -8.69927
&0.18654
&-4.02513
\\
$N_{rfb}$ & 42.69381
&18.9659
&-34.6882
\\
$K_{r}$ & 2.350899
&2.220768
&-0.31141
\\
$K_{rfb}$ & -14.5913
&-11.0303
&-15.1889
\\
$\tau_{s}$ & 0.134939
&0.212613
&0.08313
\\
$Y_{ped}$ &0
&0
&0
\\
$M_{col}$ & 0
&0
&0
\\
$A_{lat}$ & -0.09993
&-0.01799
&-0.09167
\\
$A_{lon}$ & 0.701979
&0.766162
&0.083986
\\
$B_{lat}$ & -0.07779
&0.510326
&-0.15717
\\
$B_{lon}$ & -0.09942
&-0.0212
&-0.09096
\\
$Z_{col}$ & -6.05944
&40.69521
&-18.7476
\\
$N_{ped}$ & -27.4672
&4.452297
&-36.3945
\\
$N_{col}$ & -3.22316
&3.742138
&-2.04491
\\
$C_{lon}$ & -0.09815
&0.495883
&-0.09044
\\
$D_{lat}$ & 0.793573
&0.671539
&0.025611
\\
  \hline
  \end{tabular}

\begin{figure}[htp]
\centering
\includegraphics[clip,width=0.7\columnwidth]{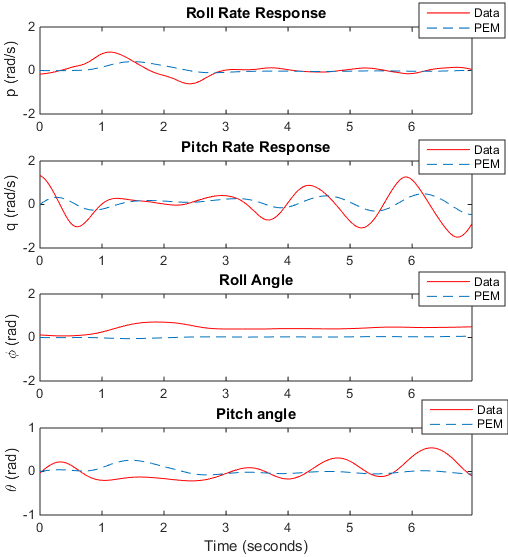}%
\captionof{figure}{\textbf{Time domain verification of validation data using PEM}}
\includegraphics[clip,width=0.7\columnwidth]{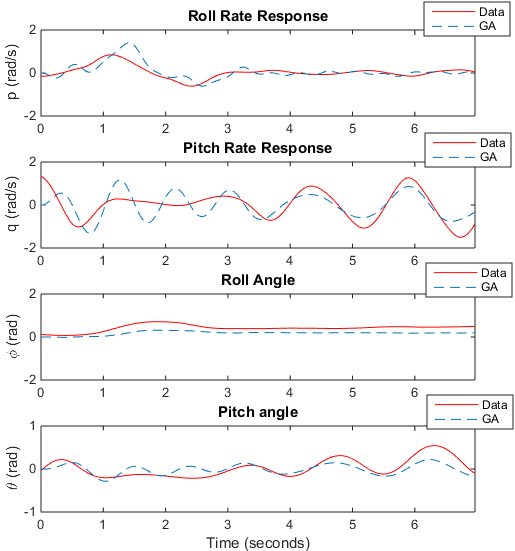}%
\captionof{figure}{\textbf {Time domain verification of validation data using GA}}

 \includegraphics[clip,width=0.7\columnwidth]{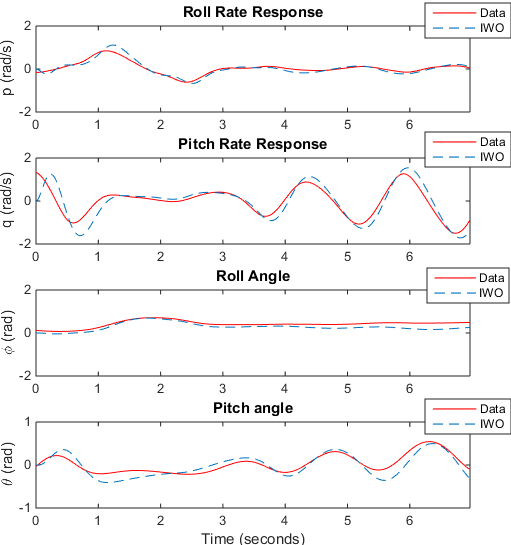}%
 \captionof{figure}{\textbf{Time domain verification of validation data using IWO algorithm}}
\end{figure}

\section{conclusions}
\justify
A highly accurate state space model of the helicopter dynamics was identified using the IWO algorithm. From this we can conclude that the IWO algorithm is a powerful tool for identifying helicopter dynamics and possibly even other complex systems. This work could be taken further by using the identified model to design controllers to enable helicopter autonomy. Apart from that, there is a big scope in comparing the effectiveness of metaheuristic algorithms on state space and other nonlinear models like Nonlinear Autoregressive with Exogenous inputs (NARX) and   Hammerstein-Wiener to have a better understanding of the algorithms. 

\begin{flushleft}
Author Contact : \newline
Navaneethkrishnan B, navaneeth@aero.iisc.ernet.in  \newline Pranjal Biswas, pranjalbiswas27@gmail.com \newline Saumya Kumaar, saumya@aero.iisc.ernet.in \newline Gautham Anand, gauthama.anand@gmail.com  \newline S N Omkar, omkar@aero.iisc.ernet.in
\end{flushleft}

\section{References} 

$^{1}$Valavanis, Kimon P., ed. \textit{Advances in unmanned aerial vehicles: state of the art and the road to autonomy} Vol. 33. Springer Science \&Business Media, Berlin, Germany, 2008, Chapter 1.\\

$^{2}$Wong, K.C., "Survey of Regional Developments: Civil Applications,"\textit{ UAV Australia Conference}, Melbourne, Australia, 8-9 February 2001 pp. 8-9.\\

$^{3}$R. Sugiura, N. Noguchi, and K. Ishii, "Remote-sensing technology for vegetation monitoring using an unmanned helicopter," \textit{Biosyst. Eng.},vol. 90, no. 4, pp. 369–379, Apr. 2005.\\

$^{4}$M. Bernard, K. Kondak, I. Maza, and A. Ollero, "Autonomous transportation and deployment with aerial robots for search and rescue missions," \textit{Journal of Field Robotics}, vol. 28, no. 6, pp. 914-931, 2011.\\

$^{5}$Bart Theys, “Which one has the highest range: VTOL, helicopter, multi-rotor (or flying squirrel?)”, \textit{DIY Drones}, accessed 25th May 2017,$<$http://diydrones.com/profiles/blogs/what-has-the-best-range-vtol-helicopter-multi-rotor-or-flying$>$ \\

$^{6}$Mettler Bernard, Takeo Kanade, and Mark Brian Tischler.\textit{ System identification modeling of a model-scale helicopter}. Carnegie Mellon University, The Robotics Institute, 2000.\\

$^{7}$Mettler Bernard, \textit{Identification modeling and characteristics of miniature rotorcraft}. Springer Science \& Business Media, Berlin, Germany, 2013.\\

$^{8}$Tischler M.B., Remple R.K., "Aircraft and Rotorcraft System Identification," \textit{Engineering Methods with Flight Test Examples}, American Institute of Aeronautics and Astronautics, Virginia, USA, 2006.\\

$^{9}$Shim, D. Hyunchul, Hyoun Jin Kim, and Shankar Sastry. "Control system design for rotorcraft-based unmanned aerial vehicles using time-domain system identification." In\textit{ Control Applications, 2000. Proceedings of the 2000 IEEE International Conference} on, pp. 808-813. IEEE, 2000.\\

$^{10}$Li, Ping, Ian Postlethwaite, and Matthew C. Turner. "Subspace-based System Identification for Helicopter Dynamic Modelling." \textit{American Helicopter Society 63rd Annual Forum}, Virginia Beach, VA, May 1-3, 2007.\\

$^{11}$Raptis, Ioannis A., Kimon P. Valavanis, and Wilfrido A. Moreno, "System identification and discrete nonlinear control of miniature helicopters using backstepping," \textit{Journal of intelligent \& robotic systems}, vol. 55, no. 2, pp. 223-243, 2009.\\

$^{12}$Raptis, Ioannis A., Kimon P. Valavanis, and Wilfrido A. Moreno, "System Identification for a Miniature Helicopter at Hover Using Fuzzy Models”, \textit{Journal of intelligent \& robotic systems}, vol. 55, no. 2, pp. 347-360, 2009.\\

$^{13}$Khajehzadeh M, Taha MR, El-Shafie A, Eslami M, "A survey on metaheuristic global optimization algorithms," \textit{Research Journal of Applied Sciences, Engineering and Technology}, vol. 3, no. 6, pp. 569-78, 2011.\\

$^{14}$Tang, Shuai, Xinye Zhao, Li Zhang, and ZhiQiang Zheng. "System identification of heave-yaw dynamics for small-scale unmanned helicopter using genetic algorithm." In \textit{32nd Chinese Control Conference} (CCC), pp. 1797-1802. IEEE, 2013\\

$^{15}$Mehrabian, A.R. and Lucas, C., 2006. A novel numerical optimization algorithm inspired from weed colonization.\textit{ Ecological informatics}, 1(4), pp.355-366.\\

$^{16}$Giri, Ritwik, Aritra Chowdhury, Arnob Ghosh, Swagatam Das, Ajith Abraham, and Vaclav Snasel. "A modified invasive weed optimization algorithm for training of feed-forward neural networks." \textit{IEEE International Conference on Systems Man and Cybernetics (SMC)}, 2010, pp. 3166-3173. IEEE, 2010.\\

$^{17}$Zhao, Huan, Pei-hong Wang, Xianyong Peng, Jin Qian, and Quan Wang. "Constrained optimization of combustion at a coal-fired utility boiler using hybrid particle swarm optimization with invasive weed." In \textit{Energy and Environment Technology}, 2009. ICEET'09. International Conference on, vol. 1, pp. 564-567. IEEE, 2009.\\

$^{18}$Kundu, Debarati, Kaushik Suresh, Sayan Ghosh, and Swagatam Das. "Designing Fractional-order PI λ D μ controller using a modified invasive Weed Optimization algortihm." In \textit{World Congress on Nature \& Biologically Inspired Computing}, 2009. NaBIC 2009., pp. 1315-1320. IEEE, 2009.\\

$^{19}$Pahlavani, P., Delavar, M.R. and Frank, A.U., 2012. Using a modified invasive weed optimization algorithm for a personalized urban multi-criteria path optimization problem.\textit{ International Journal of Applied Earth Observation and Geoinformation}, 18, pp.313-328.\\

$^{20}$Chen, Z., Wang, S., Deng, Z. and Zhang, X., 2011, September. “Tuning of auto-disturbance rejection controller based on the invasive weed optimization”.\textit{ Sixth International Conference on Bio-Inspired Computing: Theories and Applications (BIC-TA)}, 2011 (pp. 314-318). IEEE.\\

$^{21}$Tischler Mark B., Jay W. Fletcher, Vernon L. Diekmann, Robert A. Williams, and Randall W. Cason. \textit{Demonstration of frequency-sweep testing technique using a Bell 214-ST helicopter}. No. NASA-A-87073. National Aeronautics and Space Administration, Moffett Feild CA Ames Research Centre, 1987.\\


\end{document}